\documentclass[letterpaper, 10 pt, conference]{ieeeconf}
\IEEEoverridecommandlockouts
\usepackage{cite}
\usepackage{amsmath,amssymb,amsfonts}

\usepackage{fancyhdr}
\usepackage{algorithm} % added for pseudo algorithm
\usepackage{booktabs} % added
\usepackage{subfigure} % added
\usepackage{etoolbox}
\makeatletter
\patchcmd{\@makecaption}
  {\scshape}
  {}
  {}
  {}
\makeatletter
\patchcmd{\@makecaption}
  {\\}
  {.\ }
  {}
  {}
\makeatother

\usepackage{caption}
\newcommand{\Rs}{R_1}
\newcommand{\Rc}{R_2}

\usepackage{algorithmic}
\usepackage{graphicx}
\usepackage{textcomp}
\usepackage{xcolor}
\def\BibTeX{{\rm B\kern-.05em{\sc i\kern-.025em b}\kern-.08em
    T\kern-.1667em\lower.7ex\hbox{E}\kern-.125emX}}
\begin{document}

\title{\LARGE \bf Suicidal Pedestrian: Generation of Safety-Critical Scenarios for Autonomous Vehicles
\thanks{The work is supported by FCAI (Finnish Center for Artificial Intelligence)}
}

% PINs
% Yuhang: 63093
% Kalle: 63079
% Amin: 63113
% Joni: 63098
% Alex: 63089

% \author{\IEEEauthorblockN{Yuhang Yang}
% \IEEEauthorblockA{\textit{Departement of Computer Science} \\
% \textit{Aalto University}\\
% Espoo, Finland \\
% yuhang.yang@aalto.fi}
% \and
% \IEEEauthorblockN{Kalle Kujanpää}
% \IEEEauthorblockA{\textit{Departement of Computer Science} \\
% \textit{Aalto University}\\
% Espoo, Finland \\
% kalle.kujanpaa@aalto.fi}
% \and
% \IEEEauthorblockN{Amin Babadi}
% \and
% \IEEEauthorblockN{Joni Pajarinen}
% \IEEEauthorblockA{\textit{Department of Electrical Engineering and Automation} \\
% \textit{Aalto University}\\
% Espoo, Finland \\
% joni.pajarinen@aalto.fi}
% \and
% \IEEEauthorblockN{Alexander Ilin}
% \IEEEauthorblockA{\textit{Departement of Computer Science} \\
% \textit{Aalto University}\\
% Espoo, Finland \\
% alexander.ilin@aalto.fi}
% }

\author{Yuhang Yang$^{1}$, Kalle Kujanpää$^{2}$, Amin Babadi$^{3}$, Joni Pajarinen$^{1}$, and Alexander Ilin$^{2}$
\thanks{$^{1}$Yuhang Yang and Joni Pajarinen are with the Department of Electrical Engineering and Automation, Aalto University, Espoo, Finland. {\tt\footnotesize yuhang.yang@aalto.fi}; {\tt\footnotesize joni.pajarinen@aalto.fi}}
\thanks{$^{2}$Kalle Kujanpää and Alexander Ilin are with the Department of Computer Science, Aalto University, Espoo, Finland. {\tt\footnotesize kalle.kujanpaa@aalto.fi}; {\tt\footnotesize alexander.ilin@aalto.fi}}
\thanks{$^{3}$Amin Babadi is with Bugbear Entertainment Oy. Helsinki, Finland. Work done while at Aalto University. {\tt\footnotesize amin.babadi@bugbear.fi}}
}

% For commenting
\newcommand{\kalle}[1]{\textcolor{green}{Kalle: #1}}
\newcommand{\alex}[1]{\textcolor{blue}{Alex: #1}}
\newcommand{\joni}[1]{\textcolor{red}{Joni: #1}}

\maketitle

\thispagestyle{fancy}
\fancyhead{}
\rhead{}
\lfoot{26th IEEE International Conference on Intelligent Transportation Systems (ITSC 2023), Bilbao, Bizkaia, Spain}
\cfoot{\quad}
\renewcommand{\headrulewidth}{0pt}

\pagestyle{empty}

\begin{abstract}

Developing reliable autonomous driving algorithms poses challenges in testing, particularly when it comes to safety-critical traffic scenarios involving pedestrians. An open question is how to 
simulate rare events, not necessarily found in autonomous driving datasets or scripted simulations, but which can occur in testing, and, in the end may lead to severe pedestrian related accidents.
This paper presents a method for designing a suicidal pedestrian agent within the CARLA simulator, enabling the automatic generation of traffic scenarios for testing safety of autonomous vehicles (AVs) in dangerous situations with pedestrians.
The pedestrian is modeled as a reinforcement learning (RL) agent with two custom reward functions that allow the agent to either arbitrarily or with high velocity to collide with the AV.
%Kalle's proposal
Instead of significantly constraining the initial locations and the pedestrian behavior, we allow the pedestrian and autonomous car to be placed anywhere in the environment and the pedestrian to roam freely to generate diverse scenarios.
%Original: By allowing the pedestrian to freely explore the environment while maintaining a constrained initial distance from the vehicle, the generated scenarios offer greater diversity as the pedestrian and autonomous car can be placed anywhere.
%Joni's proposal: Since we allow RL to choose pedestrian behavior and starting position from a limited distance, and the AV position, the system finds automatically interesting dangerous scenarios.
To assess the performance of the suicidal pedestrian and the target vehicle during testing, we
propose three collision-oriented evaluation metrics. Experimental results involving two state-of-the-art autonomous driving algorithms trained end-to-end with imitation learning from sensor data demonstrate the effectiveness of the suicidal pedestrian in identifying decision errors made by autonomous vehicles controlled by the algorithms. 

%state-of-the-art autonomous vehicles \joni{We should make also this more accurate similarly to the previous comment}.

%\joni{Comment: there are a lot of different kinds of autonomous driving algorithms, related to
%sensor processing, path planning, control etc. We should clearly say which level of abstraction we focus on. We are focusing on end-to-end learning? And state-of-the-art "autonomous driving algorithms" learn end-to-end driving using low level control from images?}

%\joni{Hmm, actually what we mean with "autonomous driving algorithm", is defined in Fig.1 caption. Maybe just say end-to-end autonomous driving algorithm, or, use some other specification. Some people may not check the caption.}

%\kalle{end-to-end imitation learning from sensor data}

% specifically trained using a model-free learning algorithm incorporating 

% By allowing the pedestrian to freely explore the environment while maintaining a constrained initial distance from the vehicle, the generated scenarios offer greater diversity as the pedestrian and autonomous car can be placed anywhere. 

% Additionally, three collision-oriented evaluation metrics are proposed to assess the performance of the suicidal pedestrian and the target vehicle during testing. Experimental results involving two state-of-the-art autonomous driving algorithms demonstrate the effectiveness of the suicidal pedestrian in identifying decision errors made by autonomous vehicles in pedestrian-related traffic scenarios.

\end{abstract}

% \begin{IEEEkeywords}
% %traffic scenario generation,
% autonomous vehicle testing, reinforcement learning, adversarial learning
% \end{IEEEkeywords}

\section{Introduction} 

Autonomous driving (AD) is a captivating field of research that holds great potential for enhancing household mobility, optimizing traffic efficiency, and ensuring safety. In recent years, AD has gained considerable attention, and remarkable advancements have been made. Two approaches have emerged: modular driving systems that design and train each sub-module separately according to its functions \cite{modular_survey, efficient_control_avoidance}, and end-to-end models that directly perform decisions based on raw sensor inputs \cite{learning_in_one_day,learningbycheating}. However, despite these advancements, deploying AD on a large scale remains a significant challenge. A crucial reason for this is the difficulty, danger, and time-consuming nature of testing and validating autonomous vehicles (AVs), particularly in scenarios involving pedestrian safety.

\begin{figure}[htbp]
    \centerline
    {
        \includegraphics[width=1.0\linewidth]{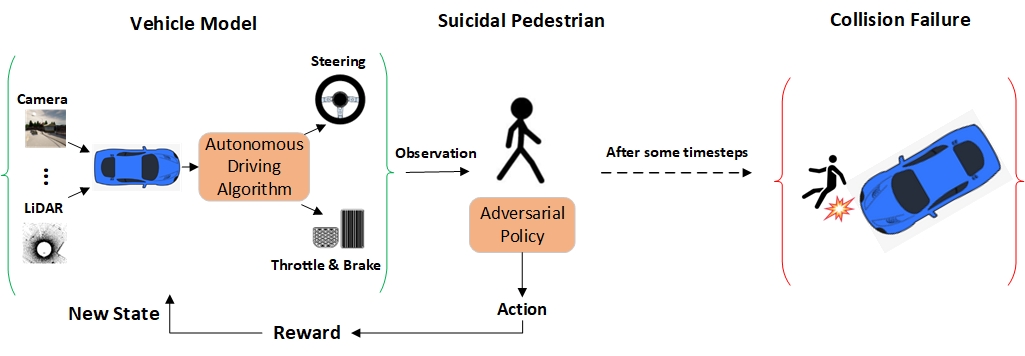}
    }
    \caption{Overview of the proposed suicidal pedestrian traffic scenario. The autonomous vehicle (AV) is controlled by an autonomous driving (AD) algorithm, which takes inputs from various sensors, producing low-level control commands to drive the vehicle safely. The pedestrian, modeled as a reinforcement learning (RL) agent, observes the location and velocity of the vehicle, and tries to hit the car with an adversarial policy learned from reward feedback.}
    \label{fig:method_framework}
\end{figure}

Several datasets \cite{vision_kitti, waymo_dataset, semantickitti} have been provided for AV testing. However, most of these manually collected data contain few safety-critical scenarios, rendering a severe overestimation of the safety performance of the testing vehicle. Other popular practices for AV testing focus on generating traffic scenarios \cite{traffic_scenario_2, traffic_scenario_3, traffic_scenario_deeproad}. While these practices enrich the range of test scenarios and expedite the validation process, they are often limited to specific scenes, such as highways or intersections, and do not adequately consider pedestrian interactions.

In this paper, we propose a method for automatically generating pedestrian-related, safety-critical traffic scenarios specifically for AV testing. By optimizing the pedestrian's behavior in the scene, we guide the pedestrian to exhibit suicidal actions with the intention of colliding with the moving car, thereby forcing the vehicle to take emergency actions. To achieve this, we formulate the suicidal pedestrian as a reinforcement learning (RL) agent and train it using a model-free RL algorithm. Additionally, to enable the pedestrian to adapt to various scenes, we design an observation space based on pedestrian characteristics and impose constraints on the initial distance between the test vehicle and the pedestrian. To demonstrate the effectiveness of our approach in generating suicidal pedestrian-based scenarios, we conduct extensive experiments in different environments, employing diverse driving policies.

The main contributions of this paper are as follows:
\begin{enumerate}
\item {Proposing a method for generating pedestrian-related safety-critical traffic scenarios dedicated to AV testing.} 
\item {Designing a suicidal pedestrian as an RL agent that aims to collide with the AV under test and training the agent using a model-free RL algorithm.}
\item {Generalizing %the behavior of
the suicidal pedestrian to test various driving policies in different environments after training it against a specific driving agent in limited situations.}
\item {Experimentally demonstrating the effectiveness of our suicidal pedestrian in identifying AV decision failures through testing with two state-of-the-art AD algorithms.}
\end{enumerate}

\section{Related work}

\subsection{Traffic Scenario Generation for Vehicle Testing}

Traffic scenario generation for vehicle testing aims to construct diverse traffic situations using simulators in order to expedite and streamline the AV testing process because real-world testing can be dangerous and expensive, particularly for safety-critical scenarios.

Recently, a lot of research has been presented on traffic scenario generation \cite{traffic_scenario_1, traffic_scenario_2, traffic_scenario_3, traffic_scenario_deeproad}. In \cite{traffic_scenario_2}, an adversarial driving scenario for AV testing is proposed, which involves training an adversarial car using Bayesian optimization and modeling the unknown cumulative performance of the test agent as a Gaussian process. Another work \cite{traffic_scenario_1} models a three-agent environment to test AVs for detecting decision errors and improve their performance through a two-step training framework. The first step involves training an adversarial vehicle to identify failures in the test cars, while the second step focuses on retraining the autonomous car based on these failure states to enhance its robustness. Furthermore, authors in \cite{traffic_scenario_3} propose an intelligent testing environment to validate the statistical capacity boundary of AVs in an accelerated mode. By removing non-safety-critical states and reconnecting critical ones, the Markov decision process (MDP) is modified to contain only relevant information, thereby densifying the training data and reducing the time required for AV testing. However, these studies pay little attention to pedestrians, limiting their applicability.

More recent works have further studied the behavior of pedestrians. In \cite{traffic_scenario_4}, pedestrians are trained to cross roads through crosswalks when a test vehicle approaches. However, the pedestrian trajectory is pre-scripted, constraining the proposed method from being generalized to other environments. Drawing inspiration from various existing pedestrian models \cite{pedestrian_model_1, traffic_scenario_simulation_6}, a pedestrian-placement model \cite{traffic_scenario_simulation_5} learns to adversarially synthesize test scenarios to increase the likelihood of collisions with a test AV given a fixed driving policy and pedestrian behavior model. However, this approach was not evaluated against state-of-the-art AD algorithms.
%a pedestrian-placement model \cite{traffic_scenario_simulation_5} is proposed to learn the optimal placement of pedestrians in a way that increases the likelihood of collisions with a test AV, rather than directly controlling their actions or trajectories. This method comprises three components. The first two components are the AV algorithm and the pedestrian behavior model, both of which are frozen and remain unchanged. The last component is the learnable adversarial test synthesizer that learns to initialize the pedestrian in an appropriate location according to the selected pedestrian behavior model, occlusions, and scene semantics.

\subsection{Reinforcement Learning}

RL algorithms guide agents to interact with the environment and to learn behaviors through a trial-and-error style without explicit human supervision. %\cite{sutton2018reinforcementbook}.
The RL problems are modeled as MDPs and the objective of RL is to maximize the rewards in the MDP by learning how to act.
%, and they are widely used to solve MDP problems. 
Specifically, for a given MDP, RL algorithms 
aim to learn an optimal policy $\pi^{*}(s)$ that maximizes the expectation of the cumulative discounted return for every state $s \in \mathcal{S}$:
\begin{equation}
    \max \limits_{\pi} \mathbb{E}\bigg [\sum^{T}_{t=0} \gamma^{t} R_{t+1} \bigg | s \bigg],
    \label{equation_Return}
\end{equation}
where $T$ is the time horizon, $\gamma$ the discount factor, and $R$ the reward function that at each step $t$ depends on the action $a_t$ taken by the policy $\pi$ in state $s_t$.
%$V$ is the value function, $\pi(s_t)$ the action chosen in state $s_t$, $R$ the reward function, $\gamma$ the discount factor, $\mathcal{S}$ the state space, and $T$ the time horizon. We use a finite $T$.
% that can maximize the expectation of cumulative discounted reward for every initial state $s_0 \in \mathcal{S}$:
% \begin{equation}
%     V_{\pi^{*}}(s_0) = \max \limits_{\pi} \mathbb{E}\bigg(\sum^{T}_{t=0} {\gamma^{t} R(s_{t+1}, s_{t}, \pi(s_t))} \bigg),
% \label{equation_Return}
% \end{equation}
%where $V$ is the value function, $\pi(s_t)$ the action chosen in state $s_t$, $R$ the reward function, $\gamma$ the discount factor, $\mathcal{S}$ the state space, and $T$ the time horizon. We use a finite $T$.

%T the finite or infinite time horizon. In our settings, T is finite.

%There exist many RL algorithms.
%Based on their methodology,
% RL algorithms can be categorized into two types: model-free algorithms and model-based algorithms. Model-free algorithms learn policies to directly maximize the rewards without an explicit model of the environment, that is they assume the environment is a black box that only produces rewards for agent actions. In contrast, model-based algorithms learn a model of the environment dynamics,
%transition and reward functions to describe the environment mechanism
%then optimize the behavior with the learned model.

In the AD area, RL algorithms have been widely used either for developing new driving systems %\cite{end_to_end_model_free_RL_driving, chekroun2021gri, chen2021worldonrail}
\cite{end_to_end_model_free_RL_driving, chen2021worldonrail} or for generating traffic scenarios \cite{adaptive_testing, adaptive_testing_augmentation, zero-shot}. %Our work is also highly related to RL. %Specifically,
We model the traffic scenario as an MDP and train our suicidal pedestrian using a model-free RL algorithm, PPO \cite{PPO}.

\section{Method}

Our work focuses on the generation of safety-critical traffic scenarios involving pedestrians to facilitate the testing of AVs in urban settings. As illustrated in Fig. \ref{fig:method_framework}, the generated scenarios contain two agents: the AV being tested and the suicidal pedestrian (Section~\ref{Method_Pedestrain_modeling}). The AV is controlled by some state-of-the-art driving algorithms, which take sensor observations as inputs and produce low-level commands such as steering angle and acceleration to ensure safe driving. On the other hand, the pedestrian, modeled as an RL agent and trained with a model-free RL algorithm (Section~\ref{Method_Policy_optimization}), observes the location and motion of the AV and attempts to collide with the car, thereby causing the AV failure. Generating the training scenarios involving the pedestrian and the target vehicle is a non-trivial process. If the pedestrian and vehicle are too close to each other, causing a collision can be very easy, and if they are far away or move in opposite directions, it is very difficult. To address this issue, we constrain the set of initial states (Section~\ref{Method_Policy_optimization}).%(Section~\ref{Method_Pedestrain_Vehicle_constraint}).

\subsection{Walking as a Markov Decision Process} \label{Method_Pedestrain_modeling}

One of the central challenges addressed in this paper is formulating the testing scenario as a Markov decision process (MDP). Given that the testing AV is already well-trained with fixed driving policies, our focus lies on modeling the pedestrian. Consequently, it becomes crucial to precisely define the state space $S$, action space $A$, and reward function $R$ for the pedestrian. The state transition dynamics are implicitly determined by the simulator once the aforementioned three elements are established.

\textit{1) State Space}: The state input for our suicidal pedestrian agent captures how the agent perceives the environment. Since the pedestrian can successfully collide with the car by knowing the vehicle's position and velocity information, a finite-dimensional vector containing this information suffices for our collision-seeking suicidal pedestrian. Additionally, we consider how to represent the position and velocity. It can either be directly represented in world coordinates 
%using an absolute form 
or in a relative form by describing it in the pedestrian coordinate system via coordinate transformation. In this paper, we adopt the relative representation due to its rotation and translation invariance, which enhances the generalization capability of our suicidal pedestrian.

Therefore, we use the following state space:
\begin{equation}
    s=[\alpha, d,\beta, v]
\end{equation}
where $\alpha$ is the angle of direction and $d$ the distance to the target vehicle from the pedestrian, $\beta$ is the relative direction in which the target vehicle is moving and $v$ is the relative scalar speed.  

\textit{2) Action Space}: The action of our suicidal pedestrian is determined by the forward direction angle and the scalar velocity. The forward direction angle, ranging from $[-\pi, \pi]$, specifies what direction the pedestrian will walk toward, while the scalar velocity ranging from [0, 3.5] in $m/s$ describes how fast the pedestrian is. Notably, since the input state is represented in the pedestrian coordinate, the output action is also represented in this coordinate. However, both the pedestrian agent and the AV move in the environment defined in the world coordinate. Therefore, the pedestrian action, especially for its forward direction angle, must be transformed back to the world coordinates. % to avoid potential errors.

\textit{3) Reward Functions}: The reward function plays an essential role in training the pedestrian policy.
%In order to illustrate that a pedestrian with a velocity-proportional and collision-part-different reward can lead to creating more complex and unpredictable adversarial behaviors than the collision-focus pedestrian agent, two different types of reward functions are designed: the constant reward function $\Rs$ and the combinational reward function $\Rc$.
We have considered two types of rewards:
\begin{itemize}
\item Reward $\Rs$ which aims to maximize the collision rate without considering the velocity of the vehicle:
\begin{equation*}
    \Rs = \left\{
    \begin{aligned}
        1, & \quad\text{if hit the vehicle} \\
        0, & \quad\text{otherwise} \\
    \end{aligned}
    \right.
\label{reward_function:constant}
\end{equation*}

\item Reward $\Rc$  which encourages the pedestrian to generate the most hazardous collisions by encouraging collisions when the vehicle is driving at high speeds:
\begin{equation*}
\Rc = \left\{
    \begin{aligned}
        \max(3, 1.5 v_c),& \ \text{if hit the front of vehicle} \\
        \max(1, 0.5 v_c),& \ \text{if hit other parts of vehicle} \\
        0, & \ \text{otherwise} \\
    \end{aligned}
    \right.
\label{reward_function:combinational}
\end{equation*}
where $v_c$ is the velocity of the vehicle when the collision happens. The shaped reward forms a natural curriculum and helps learn complex and unpredictable behaviors, such as exploiting occluded areas, required to fool the AD algorithms into dangerous frontal collisions.  
%, for example, the agent learns to use occluded areas, such as the front corners to collide with the vehicle.

\end{itemize}

\begin{table}[t]
\caption{The PPO hyperparameters}
\centering
%\renewcommand{\arraystretch}{1.2}
%\resizebox{0.45\textwidth}{!}{ %
\begin{tabular}{ll}
\hline
\textbf{Parameter} & \textbf{Value} \\ \hline
No. total training steps & 70000 \\
No. epochs when optimizing the surrogate loss & 10 \\
No. env. steps to run per update & 150 \\
Batch size & 64 \\  %\hline
Learning rate for actor and critic networks %$\lambda$
& $3 \times 10^{-4}$ \\
Discount factor %$\gamma$
& 0.98 \\
%Bias-variance trade-off factor for GAE %$\lambda_{gae}$
$\lambda$ for Generalized Advantage Estimate (GAE) 
& 0.95 \\
Objective clipping value %$\lambda_{clip}$
& 0.2 \\ %\hline
Value loss coefficient & 0.5 \\
Entropy regularization coefficient & 0.01 \\
\hline
\end{tabular}
%}%
\label{tab:PPO_train_parameter}
\end{table}

%On the other hand, $\Rc$ considers more conditions when maximizing the collision rate. First, the pedestrian should behave more positively to hitting the vehicle, i.e., it needs to collide with a vehicle driving at high speed as much as possible. Second, the pedestrian should perform complex and unpredictable behaviors to cheat the vehicle. Rather than hitting the car from the central front, it is preferred to create collisions from some view-occlusion areas, such as the front corners or sides of the vehicle. To this end, $\Rc$ is formulated as:

% Furthermore, to ensure collision diversity and complexity, the front collision part in 
% \eqref{reward_function:combinational} is intentionally designed to contain some areas from the sides. Specifically, the front collision part covers the area ranging from $[-60^{\circ}, 60^{\circ}]$ along the forward x-axis of the vehicle, with the car center as the origin.

\subsection{Policy Optimization} \label{Method_Policy_optimization}

We use a continuous-action model-free RL algorithm Proximal Policy Optimization (PPO) \cite{PPO} to train the suicidal pedestrian.
%The PPO performs online learning in each training episode and outputs a policy, i.e., a probability distribution function to represent the strategy. %Furthermore, it is a policy-based RL algorithm.
%PPO uses the policy gradients to update its current policy by increasing the probability of actions that bring higher returns. 
%strategy and utilizes the received reward to weaken or enhance the probability of chosen behaviors so that behaviors that bring higher returns are more likely to be selected.
We estimate the advantage function %for the policy updates 
with GAE \cite{gae}, and use the hyperparameters described in Table~\ref{tab:PPO_train_parameter}.

%\subsection{Pedestrian-Vehicle Initial Distance Constraint} \label{Method_Pedestrain_Vehicle_constraint}

%The pedestrian agent should be properly initialized nearby the vehicle; otherwise, it may be too trivial or too difficult for this agent to hit the vehicle.
At the beginning of each episode, we spawn the pedestrian close to the vehicle within a sector area from $-60^{\circ}$ to $60^{\circ}$ based on the forward direction of the vehicle, with the distance varying from 7m to 30m. This creates a task of a suitable difficulty level. When the distance is lower, it is easier to hit the vehicle. On the other hand, a larger distance allows the pedestrian to learn more complex behaviors, thus enhancing the diversity of the generated traffic scenarios.

\section{Experiments}

The experiments aim to demonstrate the effectiveness of our designed suicidal pedestrian used for generating safety-critical traffic scenarios for AV testing. To this end, we first train our suicidal pedestrian against one simple but effective rule-based AV. Later we evaluate the trained pedestrian in different environments to verify its ability to create collisions. Finally, we test two state-of-the-art AD algorithms with our trained suicidal pedestrian, exposing their decision errors when dealing with pedestrian-related traffic scenarios.

\begin{figure*}[t]
    \centerline
    {
        \includegraphics[width=.71\linewidth,trim={0 15mm 0 0},clip]
        {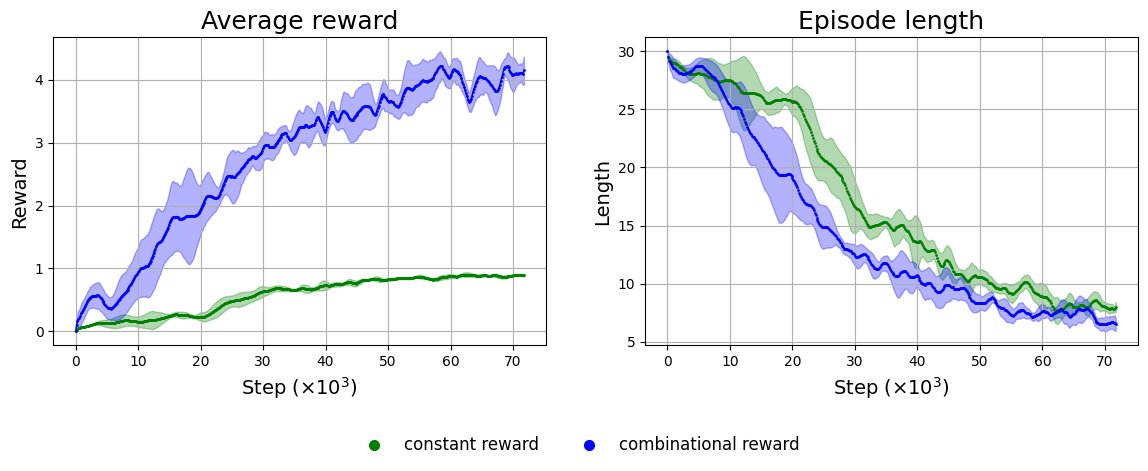}
    }
    \caption{Average rewards (left) and average episode lengths (right) during training of the suicidal pedestrian using $\Rs$ (green) and $\Rc$ (blue). The solid line represents the mean return, and the light-colored area represents the standard deviation. All plots are smoothed by the moving average over 9 data points.}
    \label{fig:train_pedestrian_plot}
\end{figure*}

\begin{table*}[h]
    \caption{Performance of the suicidal pedestrian against the CARLA AV agent.
    The best results from the point of view of the suicidal pedestrian are shown in bold.} %, and $C_S$.} %$\uparrow$ and $\downarrow$ denote that higher/lower values represent better performance.}
    \centering
    %\renewcommand{\arraystretch}{1.1}
    %\resizebox{1.0\textwidth}{!}{%
    \begin{tabular}{c|ccc|ccc}
        \hline
        & \multicolumn{3}{c}{\textbf{Train town (Town 2)}} & \multicolumn{3}{|c}{\textbf{Test town (Town 1)}} \\ \hline
        \textbf{Reward} & \textbf{Collision rate} & \textbf{Front collision rate} &  \textbf{Moving collision rate} & \textbf{Collision rate} & \textbf{Front collision rate} & \textbf{Moving collision rate} \\ \hline

        %constant reward
        $\Rs$ & $0.84 \pm 0.05$ & $0.77 \pm 0.04$ & $0.42 \pm 0.09$ & $0.79 \pm 0.00$ & $0.73 \pm 0.05$ & $0.43 \pm 0.04$ \\
        %\hline
        %combinational reward
        $\Rc$ &$\textbf{0.90} \pm 0.03$ & $\textbf{0.82} \pm 0.05$ & $\textbf{0.55} \pm 0.01$ & $\textbf{0.86} \pm 0.02$ & $\textbf{0.80} \pm 0.04$ & $\textbf{0.54} \pm 0.05$ \\

        \hline
    % \renewcommand{\arraystretch}{1.1}
    % \resizebox{1.0\textwidth}{!}{%
    % \begin{tabular}{c|cccc|cccc}
    %     \specialrule{1.5pt}{1pt}{1pt}
    %     & \multicolumn{4}{c}{\textbf{Same town (Town 2)}} & \multicolumn{4}{|c}{\textbf{New town (Town 1)}} \\ \specialrule{1.5pt}{1pt}{1pt}
    %     \textbf{Reward type} & \textbf{Collision rate ($\uparrow$)} & \textbf{Front collision ($\uparrow$)} &  \textbf{Side collision ($\downarrow$)} & \textbf{Collision running ($\uparrow$)} & \textbf{Collision rate ($\uparrow$)} & \textbf{Front collision ($\uparrow$)} &  \textbf{Side collision ($\downarrow$)} & \textbf{Collision running ($\uparrow$)}\\ \hline

    %     %constant reward
    %     $\Rs$ & $0.84 \pm 0.05$ & $0.77 \pm 0.04$ & $0.16 \pm 0.04$ & $0.42 \pm 0.09$ & $0.79 \pm 0.00$ & $0.73 \pm 0.05$ & $0.22 \pm 0.05$ & $0.43 \pm 0.04$ \\ \hline
    %     %combinational reward
    %     $\Rc$ &$\textbf{0.90} \pm 0.03$ & $\textbf{0.82} \pm 0.05$ & $\textbf{0.15} \pm 0.03$ & $\textbf{0.55} \pm 0.01$ & $\textbf{0.86} \pm 0.02$ & $\textbf{0.80} \pm 0.04$ & $\textbf{0.17} \pm 0.01$ & $\textbf{0.54} \pm 0.05$ \\

    %     \specialrule{1.5pt}{1pt}{1pt}
    \end{tabular}
    %}%
    \label{tab:pedestrian_evaluation}
\end{table*}

\subsection{Experimental Setup}

We use CARLA \cite{dosovitskiy2017carla} open-source urban driving simulator to train and validate the designed suicidal pedestrian, as well as evaluate some state-of-the-art AD algorithms.
We use Town~1 and Town~2 provided by CARLA to build our training and test environments. These towns contain T-intersections and two-lane roads.  We chose these towns because
%The motivation behind this town choice is that 
T-intersections can provide more complex traffic scenarios and two-lane roads are the main road structure in residential areas where pedestrians are more likely to appear.
We train our suicidal pedestrian against the default CARLA AV (behavior agent) with the two different reward functions, $\Rs$ and $\Rc$, and perform three training runs.

We set the episode length to 600 timesteps and run the simulator at a speed of 20 timesteps per second. This means each episode lasts 30s unless the suicidal pedestrian collides with the vehicle. Moreover, considering the speed difference, each control command for the pedestrian is repeated for 20 timesteps, equal of 1s of simulation. At the same time, we update the command for AV every timestep to avoid accidents caused by delayed controls. 

We use the OpenAI Gym \cite{OpenAIGym} framework to wrap up our designed suicidal pedestrian. We train the pedestrian with the PPO \cite{PPO} implementation from stable-baselines3 \cite{stable-baselines3}.
%Furthermore, the PPO \cite{PPO} implementation from the stable-baselines3 \cite{stable-baselines3} toolkit is selected for training our pedestrian agent.

%\subsection{Evaluation Metrics}

We evaluate the performance using the following metrics:
\begin{itemize}
\item \textit{Collision rate}: % with the target vehicle }
an overall performance metric which specifies how often the pedestrian can result in a collision with the target vehicle.

\item \textit{Moving collision rate}: collision rate when the target vehicle is moving.

\item \textit{Front collision rate}: collision rate with the front part of the target vehicle.
%\item { $C_{S}$: relative collision rate with the sides of the target vehicle }
\end{itemize}

%$C_F$ and $C_S$ are detailed metrics providing the distribution of the collision areas. Notably, different from $C_V$ and $C_R$ which are absolute collision rates calculated based on all test episodes, $C_F$ and $C_S$ are relative collision rates calculated over episodes that the collision happens.

\begin{figure*}[t]
    \centering
    \includegraphics[width=0.77\textwidth]{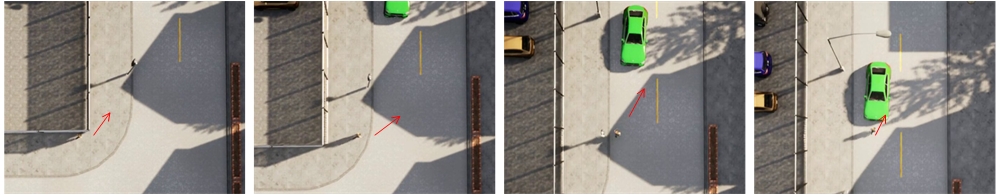}
    %\hfil
    \\
    \includegraphics[width=0.77\textwidth]{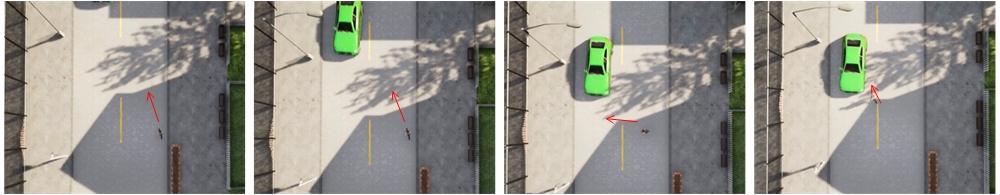}
    \caption{Typical behaviors of the pedestrian trained with $\Rc$. Red arrows represent the pedestrian direction. Top row: The pedestrian directly hits the vehicle from the central front, as the car fails to predict the pedestrian's movement. Bottom row: The pedestrian crashes into the car from the side.} %Both behaviors displayed in the top and bottom rows are desired because these behaviors would challenge the ability of AVs facing such sudden emergence.}
    \label{fig:Pedestrian_behavior_visualization}
\end{figure*}

\subsection{Training Against the CARLA AV Agent}

%\textit{1) Training the suicidal pedestrian}:  %and each reward function is trained three times to avoid occasional cases in Town 2. 
% Intentionally, we select the following locations in Town 2 to reproduce the training procedure:
% \begin{itemize}
% \item { location 1: [104.3, 241.3, 0.5],  orientation 1: [0, 0, 0]}
% \item { location 2: [88.8, 302.6, 0.5], orientation 2: [0, $-\pi$, 0] }
% \item { location 3: [190.0, 293.5, 0.5], orientation 3: [0, $\frac{\pi}{2}$, 0]  }
% \item { location 4: [193.8, 218.8, 0.5], orientation 4: [0,$-\frac{\pi}{2}$ 0]  }
% \end{itemize}
% where components of the location are $[x, y, z]$, and components of the orientation are
% $[pitch, yaw, roll]$. All of these locations and orientations are represented in the world
% coordinate.
Fig.~\ref{fig:train_pedestrian_plot} shows that training of the pedestrian policy converges after 70000 steps when trained using both rewards $\Rs$ and $\Rc$. %to some specific behaviors. Additionally, although $\pi_{const}$ converges faster than $\pi_{comb}$,
The value of the mean episode length indicates that using $\Rc$ results in a policy that is more aggressive in searching for and hitting the vehicle. %in terms of average steps needed to hit the vehicle.

Table~\ref{tab:pedestrian_evaluation} shows the performance of our suicidal pedestrian. One can see that both reward functions perform well in guiding the pedestrian to hit the target AV. $\Rc$ generally yields better performance than $\Rs$.
Note that more than half of all collisions happen when the AV does not stop in time, which corresponds to more hazardous scenarios. %and means that the pedestrian has successfully cheated the vehicle.
As for the collision areas, the front part of the vehicle receives almost $80\%$ of collisions. We can also see that the pedestrian agent generalizes well to a new town.
%We can also observe that the reward function $\Rc$ outperforms $\Rs$ on almost all metrics. %The following will discuss these key conclusions in detail.
The performance of both reward functions declines only slightly, by approximately $5\%$, when the suicidal pedestrian is deployed to a previously unknown environment.

Note that the collision rate does not reach 100\% and we see two reasons for that.
%There are two main failure cases:
%However, none of the methods can perfectly hit the vehicle. Most possible reasons come from two aspects.
First, the pedestrian uses a coordinate-based state and it may be blocked by environmental objects that are not included in the state.
%directly observes the location and velocity of the target AV while ignoring all environmental obstacles, rendering that the pedestrian may be blocked by environmental barriers when it tries to walk close to the car.
Second, sometimes the pedestrian fails to predict the future trajectory of the AV.
%the pedestrian and the vehicle are independent non-communicating players, rendering a very limited ability for the pedestrian to predict the future trajectory of the AV. 
In Fig.~\ref{fig:Pedestrian_behavior_visualization}, we visualize some typical behaviors of the suicidal pedestrian.

\subsection{Testing SOTA AD Algorithms with the Suicidal Pedestrian} %failure of autonomous vehicles}

We test two state-of-the-art AD algorithms with our suicidal pedestrian, LAV \cite{chen2022learningfromallvehicles} that plans using predicted future trajectories for all traffic participants, and InterFuser \cite{shao2023safetysafetyenhanced} that has a safety controller relying on a predicted object density map to avoid collisions.
%in terms of three metrics: the pedestrian mean episode reward, the collision rate $C_V$, and the moving collision rate $C_M$. Notably, since the evaluation entity is now the AV agent, $C_V$ and $C_M$ are interpreted from the perspective of the vehicle. For $C_V$, a value closer to 0 means the vehicle has enough capacity to avoid collisions with the suicidal pedestrian, whereas a value closer to 1 means the opposite. For $C_M$, lower values suggest that the vehicle manages to avoid collision hazards by braking, thus proving the capacity of the AV to deal with the sudden emergence of pedestrians by performing hard brakes. Furthermore, the pedestrian mean episode return is a general metric to describe the collisions, with lower values implying that it is more difficult to result in crashes or cause severe consequences. This metric is affected by both the collision rate and the collision running rate.
Note that the pedestrian was trained against the CARLA behavior agent, and it is evaluated with LAV and InterFuser without any adaptations.

\begin{table*}[tp]
    \caption{Evaluation results of two state-of-the-art AD algorithms using the suicidal pedestrian trained with reward $\Rc$.
    The best results from the point of view of the driving policy are shown in bold.}
    \centering
    %\renewcommand{\arraystretch}{1.1}
    %\resizebox{1.0\textwidth}{!}{%
    \begin{tabular}{c|ccc|ccc}
        \hline
        & \multicolumn{3}{c}{\textbf{Train town (Town 2)}} & \multicolumn{3}{|c}{\textbf{Test town (Town 1)}} \\ \hline
        \textbf{Method} & \textbf{Pedestrian reward} & \textbf{Collision rate} & \textbf{Moving collision rate} & \textbf{Pedestrian reward} & \textbf{Collision rate} & \textbf{Moving collision rate} \\ \hline
        %&  $\% \downarrow$ & $\% \downarrow$ & $\% \downarrow$ & $\% \downarrow$ & $\% \downarrow$ & $\% \downarrow$ \\ \hline
        CARLA behavior & $4.57 \pm 0.15$ & $\textbf{0.90} \pm 0.03$ & $0.55 \pm 0.02$ & $4.29 \pm 0.29$ & $\textbf{0.86} \pm 0.02$ & $0.54 \pm 0.05$ \\ %\hline
        LAV \cite{chen2022learningfromallvehicles} & $\textbf{3.56} \pm 0.26$ & $0.93 \pm 0.02$ & $0.47 \pm 0.04$ & $3.94 \pm 0.26$ & $0.90 \pm 0.03$ & $0.61 \pm 0.01$ \\
        %\hline
        InterFuser \cite{shao2023safetysafetyenhanced} & $3.77 \pm 0.38$ & $0.94 \pm 0.02$ & $\textbf{0.32} \pm 0.06$ & $\textbf{3.82} \pm 0.16$ & $0.92 \pm 0.02$ & $\textbf{0.37} \pm 0.02$ \\ 
        \hline
    \end{tabular}
    %}%
    \label{tab:vehicle_test}
\end{table*}

\begin{figure*}[tp]
    \centering
    \includegraphics[width=.42\linewidth]{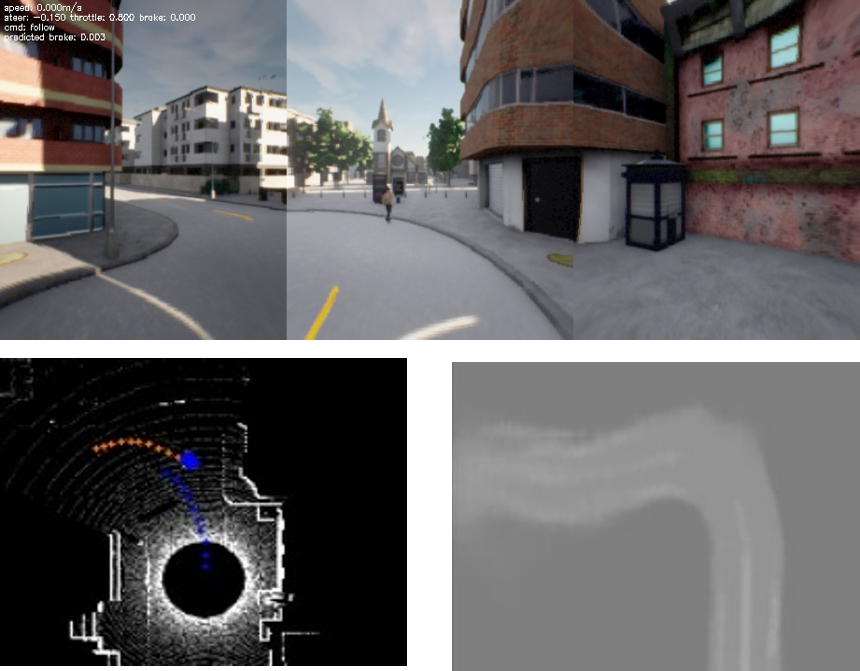}
    \hfil
    \includegraphics[width=.42\linewidth]{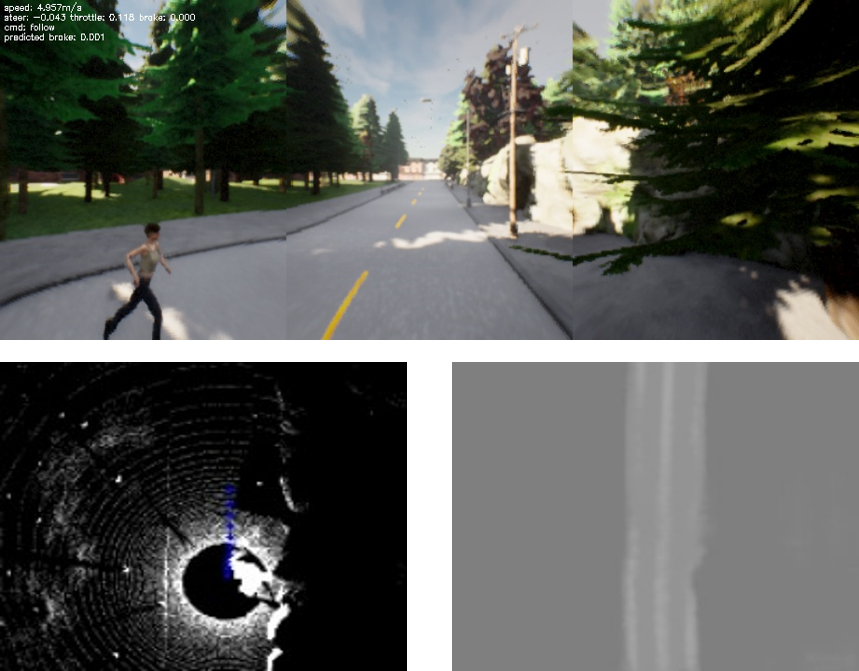}
    \caption{Visualization of two collision episodes with LAV. We present three (concatenated) camera images (top), detection and motion predictions (bottom left), and predicted road geometries (bottom right) for the two episodes. Left: LAV detects the pedestrian as a vehicle. Right: LAV fails to find the pedestrian due to insufficient fusion of images. 
    %(a) Visualization of incorrect detection error that predicts the pedestrian as a vehicle. (b) Visualization of unsuccessful detection error that fails to find the pedestrian due to insufficient fusion of images.
    }
    \label{fig:LAV_error_visualization}
\end{figure*}

\begin{figure*}[tbp]
    \centering
    \includegraphics[width=.3\textwidth]{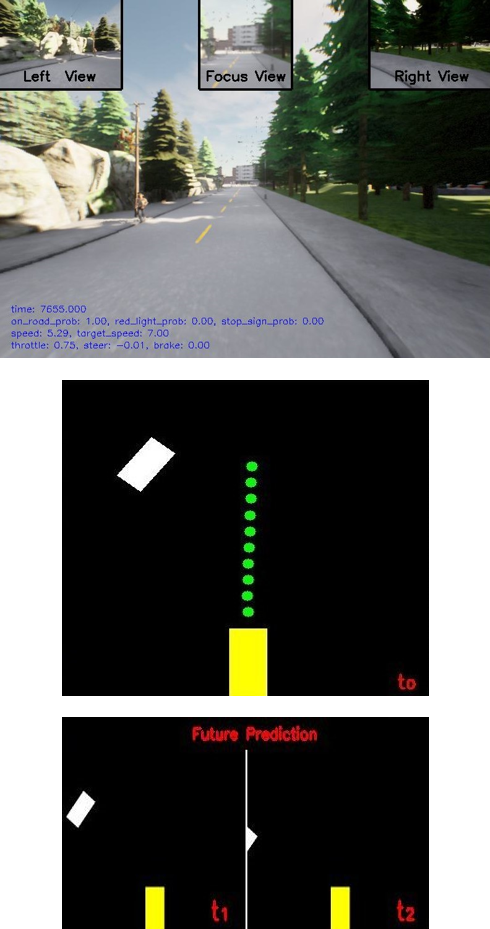}
    \hfil
    \includegraphics[width=.3\textwidth]{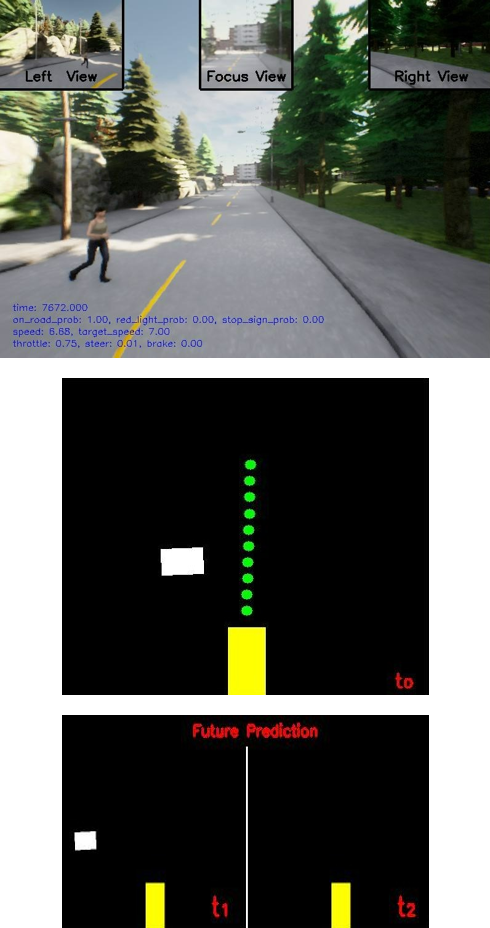}
    \hfil
    \includegraphics[width=.3\textwidth]{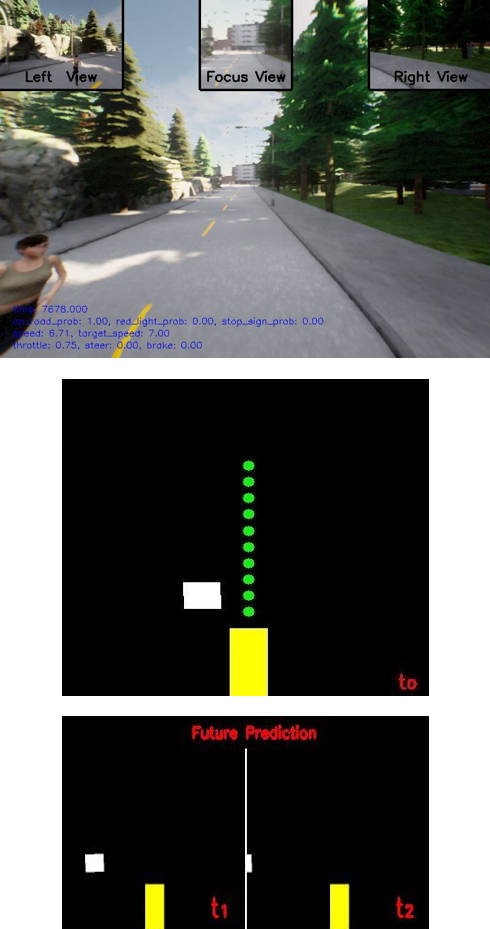}
    \caption{Visualization of a failure case of InterFuser in which the AV does not perform any actions to avoid collisions due to failing to predict the trajectory of the pedestrian. We present camera images (top row), detected traffic scenes at the current timestep (middle row) and predicted traffic scenes at the next two timesteps (bottom row). The yellow rectangle in the last two rows represents the ego vehicle, while white rectangles represent other detected objects. Green dots are the future trajectory of the ego vehicle.}
    \label{fig:InterFuser_error_visualization}
\end{figure*}

Table~\ref{tab:vehicle_test} describes the performance of LAV %\cite{chen2022learningfromallvehicles} 
and InterFuser %\cite{shao2023safetysafetyenhanced}
against the suicidal pedestrian. %
%Notably, these quantitative results of the three driving models are not comparable, i.e., it cannot be concluded that LAV and InterFuser perform better than CARLA behavior even though they achieve the lowest pedestrian reward and collision running rate, because of that the pedestrian is not trained against LAV or InterFuser. However, some helpful information can still be obtained. 
The results show that
the pedestrian can generate collisions both with LAV and InterFuser, showing potential weaknesses of these two driving algorithms. InterFuser has a much lower moving collision rate than LAV and CARLA, which suggests  
%Moreover, the difference between $C_V$ and $C_R$ of LAV is larger than that of InterFuser, which suggests
most crashes of InterFuser are not severe, while LAV is more likely to cause hazardous consequences when a collision happens.

We visualize some failures of LAV and InterFuser when dealing with our suicidal pedestrian to understand their weaknesses better. Fig.~\ref{fig:LAV_error_visualization} illustrates two typical errors of LAV: incorrect detection and unsuccessful detection. In incorrect detection, LAV detects the pedestrian as a vehicle, thus applying unreasonable dynamic models to the pedestrian to predict the corresponding trajectories. In unsuccessful detection, LAV fails to detect the pedestrian due to vision failure.
%technical flaws.
Interestingly, both errors can happen at different stages of one episode. Fig.~\ref{fig:InterFuser_error_visualization} illustrates a typical failure case of InterFuser, in which the vehicle does not perform any actions to avoid collisions with the suicidal pedestrian even if the pedestrian is detected. This failure suggests that Interfuser performs well in detection, but potential improvements should be applied to its prediction and decision-making modules.

\section{Discussion \& Future work}

This paper proposes a suicidal pedestrian model to generate safety-critical traffic scenarios for AD testing. We model the pedestrian as an RL agent and train it using a model-free PPO algorithm. Furthermore, we perform %extensive 
experiments to validate its effectiveness in generating collision scenarios. Finally, testing results of two state-of-the-art AD algorithms illustrate our suicidal pedestrian can significantly help find driving algorithm decision errors.

Our work can be extended to having more pedestrians and cars in the simulations. %, which could be more challenging for the AVs.
Another direction would be to consider different goal-conditioned pedestrians to generate more varying behaviors to address the limitation of only using the suicidal pedestrian with limited behavior diversity.
%in which the suicidal pedestrian should distinguish between the target vehicle and other vehicles, as well as pedestrians. 
Moreover, we can augment our state representation with the locations of other objects, or we could use image inputs or object-based representations to replace the hand-crafted state vector, thus allowing the pedestrian to plan movements according to the surroundings, avoid obstacles, or take advantage of obstacles to surprise the drivers. %Furthermore, combining our work with an adversarial test synthesizer proposed in prior work is also an intriguing alternative.

\bibliographystyle{IEEEtran} 
\bibliography{reference}

\begin{thebibliography}{10}
\providecommand{\url}[1]{#1}
\csname url@samestyle\endcsname
\providecommand{\newblock}{\relax}
\providecommand{\bibinfo}[2]{#2}
\providecommand{\BIBentrySTDinterwordspacing}{\spaceskip=0pt\relax}
\providecommand{\BIBentryALTinterwordstretchfactor}{4}
\providecommand{\BIBentryALTinterwordspacing}{\spaceskip=\fontdimen2\font plus
\BIBentryALTinterwordstretchfactor\fontdimen3\font minus
  \fontdimen4\font\relax}
\providecommand{\BIBforeignlanguage}[2]{{%
\expandafter\ifx\csname l@#1\endcsname\relax
\typeout{** WARNING: IEEEtran.bst: No hyphenation pattern has been}%
\typeout{** loaded for the language `#1'. Using the pattern for}%
\typeout{** the default language instead.}%
\else
\language=\csname l@#1\endcsname
\fi
#2}}
\providecommand{\BIBdecl}{\relax}
\BIBdecl

\bibitem{modular_survey}
B.~Paden, M.~{\v{C}}{\'a}p, S.~Z. Yong, D.~Yershov, and E.~Frazzoli, ``A survey
  of motion planning and control techniques for self-driving urban vehicles,''
  \emph{IEEE Transactions on intelligent vehicles}, vol.~1, no.~1, pp. 33--55,
  2016.

\bibitem{efficient_control_avoidance}
A.~Colombo and D.~Del~Vecchio, ``Efficient algorithms for collision avoidance
  at intersections,'' in \emph{Proceedings of the 15th ACM international
  conference on Hybrid Systems: Computation and Control}, 2012, pp. 145--154.

\bibitem{learning_in_one_day}
A.~Kendall, J.~Hawke, D.~Janz, P.~Mazur, D.~Reda, J.-M. Allen, V.-D. Lam,
  A.~Bewley, and A.~Shah, ``Learning to drive in a day,'' in \emph{Proceedings
  of the 2019 International Conference on Robotics and Automation
  (ICRA)}.\hskip 1em plus 0.5em minus 0.4em\relax IEEE, 2019, pp. 8248--8254.

\bibitem{learningbycheating}
D.~Chen, B.~Zhou, V.~Koltun, and P.~Kr{\"a}henb{\"u}hl, ``Learning by
  cheating,'' in \emph{Proceedings of the Conference on Robot Learning
  (CoRL)}.\hskip 1em plus 0.5em minus 0.4em\relax PMLR, 2020, pp. 66--75.

\bibitem{vision_kitti}
A.~Geiger, P.~Lenz, C.~Stiller, and R.~Urtasun, ``Vision meets robotics: The
  kitti dataset,'' \emph{The International Journal of Robotics Research},
  vol.~32, no.~11, pp. 1231--1237, 2013.

\bibitem{waymo_dataset}
P.~Sun, H.~Kretzschmar, X.~Dotiwalla, A.~Chouard, V.~Patnaik, P.~Tsui, J.~Guo,
  Y.~Zhou, Y.~Chai, B.~Caine \emph{et~al.}, ``Scalability in perception for
  autonomous driving: Waymo open dataset,'' in \emph{Proceedings of the
  IEEE/CVF conference on computer vision and pattern recognition}, 2020, pp.
  2446--2454.

\bibitem{semantickitti}
J.~Behley, M.~Garbade, A.~Milioto, J.~Quenzel, S.~Behnke, C.~Stachniss, and
  J.~Gall, ``Semantickitti: A dataset for semantic scene understanding of lidar
  sequences,'' in \emph{Proceedings of the IEEE/CVF international conference on
  computer vision}, 2019, pp. 9297--9307.

\bibitem{traffic_scenario_2}
Y.~Abeysirigoonawardena, F.~Shkurti, and G.~Dudek, ``Generating adversarial
  driving scenarios in high-fidelity simulators,'' in \emph{Proceedings of the
  2019 International Conference on Robotics and Automation (ICRA)}.\hskip 1em
  plus 0.5em minus 0.4em\relax IEEE, 2019, pp. 8271--8277.

\bibitem{traffic_scenario_3}
S.~Feng, H.~Sun, X.~Yan, H.~Zhu, Z.~Zou, S.~Shen, and H.~X. Liu, ``Dense
  reinforcement learning for safety validation of autonomous vehicles,''
  \emph{Nature}, vol. 615, no. 7953, pp. 620--627, 2023.

\bibitem{traffic_scenario_deeproad}
M.~Zhang, Y.~Zhang, L.~Zhang, C.~Liu, and S.~Khurshid, ``Deeproad: Gan-based
  metamorphic testing and input validation framework for autonomous driving
  systems,'' in \emph{Proceedings of the 33rd ACM/IEEE International Conference
  on Automated Software Engineering}, 2018, pp. 132--142.

\bibitem{traffic_scenario_1}
A.~Sharif and D.~Marijan, ``Adversarial deep reinforcement learning for
  improving the robustness of multi-agent autonomous driving policies,'' in
  \emph{Proceedings of the 2022 29th Asia-Pacific Software Engineering
  Conference (APSEC)}.\hskip 1em plus 0.5em minus 0.4em\relax IEEE, 2022, pp.
  61--70.

\bibitem{traffic_scenario_4}
D.~Karunakaran, S.~Worrall, and E.~Nebot, ``Efficient statistical validation
  with edge cases to evaluate highly automated vehicles,'' in \emph{Proceedings
  of the 2020 IEEE 23rd International Conference on Intelligent Transportation
  Systems (ITSC)}, 2020, pp. 1--8.

\bibitem{pedestrian_model_1}
A.~Rasouli and J.~K. Tsotsos, ``Autonomous vehicles that interact with
  pedestrians: A survey of theory and practice,'' \emph{IEEE Transactions on
  Intelligent Transportation Systems}, vol.~21, no.~3, pp. 900--918, 2020.

\bibitem{traffic_scenario_simulation_6}
W.~Ding, B.~Chen, M.~Xu, and D.~Zhao, ``Learning to collide: An adaptive
  safety-critical scenarios generating method,'' in \emph{Proceedings of the
  2020 IEEE/RSJ International Conference on Intelligent Robots and Systems
  (IROS)}.\hskip 1em plus 0.5em minus 0.4em\relax IEEE, 2020, pp. 2243--2250.

\bibitem{traffic_scenario_simulation_5}
M.~Priisalu, A.~Pirinen, C.~Paduraru, and C.~Sminchisescu, ``Generating
  scenarios with diverse pedestrian behaviors for autonomous vehicle testing,''
  in \emph{Proceedings of the 5th Conference on Robot Learning (CoRL)}.\hskip
  1em plus 0.5em minus 0.4em\relax PMLR, 2022, pp. 1247--1258.

\bibitem{end_to_end_model_free_RL_driving}
M.~Toromanoff, E.~Wirbel, and F.~Moutarde, ``End-to-end model-free
  reinforcement learning for urban driving using implicit affordances,'' in
  \emph{Proceedings of the IEEE/CVF Conference on Computer Vision and Pattern
  Recognition (CVPR)}, 2020.

\bibitem{chen2021worldonrail}
D.~Chen, V.~Koltun, and P.~Kr\"ahenb\"uhl, ``Learning to drive from a world on
  rails,'' in \emph{Proceedings of the IEEE/CVF International Conference on
  Computer Vision (ICCV)}, 2021, pp. 15\,590--15\,599.

\bibitem{adaptive_testing}
M.~Koren, S.~Alsaif, R.~Lee, and M.~J. Kochenderfer, ``Adaptive stress testing
  for autonomous vehicles,'' in \emph{Proceedings of the 2018 IEEE Intelligent
  Vehicles Symposium (IV)}.\hskip 1em plus 0.5em minus 0.4em\relax IEEE, 2018,
  pp. 1--7.

\bibitem{adaptive_testing_augmentation}
A.~Corso, P.~Du, K.~Driggs-Campbell, and M.~J. Kochenderfer, ``Adaptive stress
  testing with reward augmentation for autonomous vehicle validatio,'' in
  \emph{Proceedings of the 2019 IEEE Intelligent Transportation Systems
  Conference (ITSC)}.\hskip 1em plus 0.5em minus 0.4em\relax IEEE, 2019, pp.
  163--168.

\bibitem{zero-shot}
B.~Chalaki, L.~E. Beaver, B.~Remer, K.~Jang, E.~Vinitsky, A.~M. Bayen, and
  A.~A. Malikopoulos, ``Zero-shot autonomous vehicle policy transfer: From
  simulation to real-world via adversarial learning,'' in \emph{Proceedings of
  the 2020 IEEE 16th International Conference on Control \& Automation
  (ICCA)}.\hskip 1em plus 0.5em minus 0.4em\relax IEEE, 2020, pp. 35--40.

\bibitem{PPO}
\BIBentryALTinterwordspacing
J.~Schulman, F.~Wolski, P.~Dhariwal, A.~Radford, and O.~Klimov, ``Proximal
  policy optimization algorithms,'' \emph{arXiv}, 2017. [Online]. Available:
  \url{https://arxiv.org/abs/1707.06347}
\BIBentrySTDinterwordspacing

\bibitem{gae}
J.~Schulman, P.~Moritz, S.~Levine, M.~Jordan, and P.~Abbeel, ``High-dimensional
  continuous control using generalized advantage estimation,'' \emph{arXiv
  preprint arXiv:1506.02438}, 2015.

\bibitem{dosovitskiy2017carla}
A.~Dosovitskiy, G.~Ros, F.~Codevilla, A.~Lopez, and V.~Koltun, ``Carla: An open
  urban driving simulator,'' in \emph{Proceedings of the 1st Annual Conference
  on Robot Learning (CoRL)}.\hskip 1em plus 0.5em minus 0.4em\relax PMLR, 2017,
  pp. 1--16.

\bibitem{OpenAIGym}
\BIBentryALTinterwordspacing
G.~Brockman, V.~Cheung, L.~Pettersson, J.~Schneider, J.~Schulman, J.~Tang, and
  W.~Zaremba, ``Openai gym,'' \emph{arXiv}, 2016. [Online]. Available:
  \url{https://arxiv.org/abs/1606.01540}
\BIBentrySTDinterwordspacing

\bibitem{stable-baselines3}
A.~Raffin, A.~Hill, A.~Gleave, A.~Kanervisto, M.~Ernestus, and N.~Dormann,
  ``Stable-baselines3: Reliable reinforcement learning implementations,''
  \emph{Journal of Machine Learning Research}, vol.~22, no. 268, pp. 1--8,
  2021.

\bibitem{chen2022learningfromallvehicles}
D.~Chen and P.~Kr{\"a}henb{\"u}hl, ``Learning from all vehicles,'' in
  \emph{Proceedings of the IEEE/CVF Conference on Computer Vision and Pattern
  Recognition (CVPR)}, 2022, pp. 17\,222--17\,231.

\bibitem{shao2023safetysafetyenhanced}
H.~Shao, L.~Wang, R.~Chen, H.~Li, and Y.~Liu, ``Safety-enhanced autonomous
  driving using interpretable sensor fusion transformer,'' in \emph{Proceedings
  of the Conference on Robot Learning (CoRL)}.\hskip 1em plus 0.5em minus
  0.4em\relax PMLR, 2023, pp. 726--737.

\end{thebibliography}

\end{document}